\documentclass[letterpaper,10pt,conference]{ieeeconf}
\IEEEoverridecommandlockouts
\makeatletter
\let\NAT@parse\undefined
\makeatother

\usepackage{amsmath}
\usepackage{amssymb}
\usepackage{booktabs}
\usepackage{multirow}
\usepackage[font=footnotesize]{caption}
\usepackage[labelformat=simple]{subcaption}

\usepackage{tikz}
\usepackage[mode=buildnew]{standalone}
\usepackage{xspace}
\usepackage[noend]{algpseudocode}
\usepackage{tabulary}
\usepackage{verbatim}
\usepackage[numbers,sort&compress]{natbib}

\usepackage{thmtools}
\usepackage{mathtools}
\usepackage{hyperref}

\usepackage[ruled, noline]{algorithm2e}
\SetKwFor{ForPar}{for}{do in parallel}{end forpar}
\definecolor{alg}{RGB}{46, 149, 186}

\SetCommentSty{mycommfont}
\definecolor{urldarkblue}{RGB}{1, 111, 255}

\usepackage[T1]{fontenc} 
\usepackage[latin1]{inputenc}

\usepackage[english]{babel}
\usepackage[autostyle, english = american]{csquotes}
\MakeOuterQuote{"}

\usepackage[T1]{fontenc} 
\usepackage[latin1]{inputenc}
\usepackage{babel}
\usepackage{textcomp}
\usepackage{stmaryrd}
\usepackage{upgreek}
\usepackage{bm}
\usepackage{bbm}
\usepackage{url}
\usepackage{breakurl}
\usepackage{afterpage}
\usepackage{mathtools}
\usepackage[capitalise]{cleveref} 
\usepackage{pifont}
\usepackage{balance}
\hypersetup{
  bookmarksopen,
  bookmarksnumbered,
  pdfpagemode=UseOutlines,
  colorlinks=true,
  linkcolor=blue,
  anchorcolor=blue,
  citecolor=blue,
  filecolor=blue,
  menucolor=blue,
  urlcolor=blue
}

\graphicspath{{figs/}}

\newcommand{\Figref}[1]{Fig.~\ref{#1}}
\newcommand{\Tabref}[1]{Table~\ref{#1}}

\newcommand{\cmark}{\ding{51}}%
\newcommand{\xmark}{\ding{55}}%

\newcommand{\lazysp}[0]{\textsc{LazySP}\xspace}

\newcommand{\checkerboard}[0]{\textsc{Checkerboard}\xspace}
\newcommand{\intel}[0]{\textsc{Intel}\xspace}
\newcommand{\franka}[0]{\textsc{Manipulator}\xspace}

\def\hidenotes{}  
\newcommand{\bhnote}[1]{{\xxnote{BH}{Mulberry}{#1}}}
\newcommand{\rmsnote}[1]{{\xxnote{RMS}{internationalorange}{#1}}}
\newcommand{\slnote}[1]{{\xxnote{SL}{blue}{#1}}}

\newcommand{\xxnote}[3]{}
\ifx\hidenotes\undefined
  \renewcommand{\xxnote}[3]{\color{#2}{#1: #3}}
  \definecolor{SeaGreen}{HTML}{3FBC9D}
  \definecolor{Orange}{HTML}{D95F02}
  \definecolor{RacingGreen}{HTML}{004225}
  \definecolor{Mulberry}{HTML}{A93C93}
  \definecolor{internationalorange}{rgb}{1.0, 0.31, 0.0}
\fi

\newcommand{\norm}[1]{\left\lVert#1\right\rVert}

\DeclareMathOperator*{\argmax}{arg\,max}
\DeclareMathOperator*{\argmin}{arg\,min}

\newcommand{\expect}[2]{\mathbb{E}_{#1}\left[ #2 \right]}

\newcommand{\eg}{\textit{e.g.}}
\newcommand{\ie}{\textit{i.e.}}
\newcolumntype{L}[1]{>{\raggedright\arraybackslash}p{#1}}
\newcolumntype{C}[1]{>{\centering\arraybackslash}p{#1}}
\newcolumntype{R}[1]{>{\raggedleft\arraybackslash}p{#1}}

\def\real{\mathbb{R}}

\def\configspace{\mathcal{X}}
\def\config{\mathbf{x}}
\def\configs{X}
\def\state{\mathbf{x}}

\def\safeset{\mathcal{S}}
\def\thresh{\beta}
\def\conrad{\rho}
\def\J\mathbf{J}
\def\traj{\tau}

\def\lvar{z}  
\def\lvarv{\mathbf{z}}  

\def\x{\mathbf{x}}

\def\weights{\mathbf{w}}
\def\feats{\Phi}

\def\map{\Theta}
\newcommand{\probc}[2]{p(#1\,|\,#2)}
\newcommand{\prob}[1]{p(#1)}
\DeclarePairedDelimiterX{\infdivx}[2]{(}{)}{%
  #1\;\delimsize\|\;#2%
}
\newcommand{\dkldiv}{D_{KL}\infdivx}

\newcommand{\hide}[1]{}

\title{\LARGE \bf
  Stein Variational Probabilistic Roadmaps
}

\author{
  Alexander Lambert, Brian Hou, Rosario Scalise, Siddhartha S. Srinivasa, and Byron Boots%
   \thanks{
     This work was (partially) funded by the National Institute of Health R01 (\#R01EB019335),
     National Science Foundation CPS (\#1544797), National Science Foundation NRI (\#1637748), the Army Research Laboratory (\#W911NF-16-2-0008),
     the Office of Naval Research, the RCTA, Amazon, and Honda Research Institute USA.
     Brian Hou is partially supported by a NASA Space Technology Research Fellowship.}%
   \thanks{
     All authors are with the
     Paul G. Allen School of Computer Science \& Engineering,
     University of Washington, Seattle, WA 98195
     \texttt{\{lambert6, bhou, rosario, siddh, bboots\}@cs.uw.edu}.
     Rosario Scalise is also an Oak Ridge Associated University Fellow with DEVCOM Army Research Laboratory.
     The views and conclusions contained in this document are those of the authors and should not be interpreted as representing the official policies, either expressed or implied, of the Army Research Laboratory or the U.S. Government.}%
}

\begin{document}

\maketitle
\thispagestyle{empty}
\pagestyle{empty}

\begin{abstract}
Efficient and reliable generation of global path plans are necessary for safe execution and deployment of autonomous systems. In order to generate planning graphs which adequately resolve the topology of a given environment, many sampling-based motion planners resort to coarse, heuristically-driven strategies which often fail to generalize to new and varied surroundings. Further, many of these approaches are not designed to contend with partial-observability. We posit that such uncertainty in environment geometry can, in fact, help \textit{drive} the sampling process in generating feasible, and probabilistically-safe planning graphs. We propose a method for Probabilistic Roadmaps which relies on particle-based Variational Inference to efficiently cover the posterior distribution over feasible regions in configuration space. Our approach, Stein Variational Probabilistic Roadmap (SV-PRM), results in sample-efficient generation of planning-graphs and large improvements over traditional sampling approaches. We demonstrate the approach on a variety of challenging planning problems, including real-world probabilistic occupancy maps and high-dof manipulation problems common in robotics. Video, additional material and results can be found here: \url{https://sites.google.com/view/stein-prm}.
\end{abstract}
\section{Introduction and Related Work}
\label{sec:related-work}



Probabilistic roadmaps approximate a continuous configuration space with a discrete set of sampled configurations~\cite{kavraki1996prm,karaman2011sbmp}.
Theoretical bounds have been developed for the number of uniform or low-discrepancy samples required to ensure near-optimal paths~\cite{karaman2011sbmp,solovey2018rgg,janson2018deterministic}.
However, such bounds require many samples to generate a path through challenging ``narrow passage'' problems~\cite{hsu1997expansive},
yielding dense roadmaps that cannot be searched efficiently at query time.

\begin{figure}[!t]
  \begin{subfigure}[b]{0.47\linewidth}
    \centering
    \includegraphics[trim={8 8 8 8},clip,height=10\baselineskip]{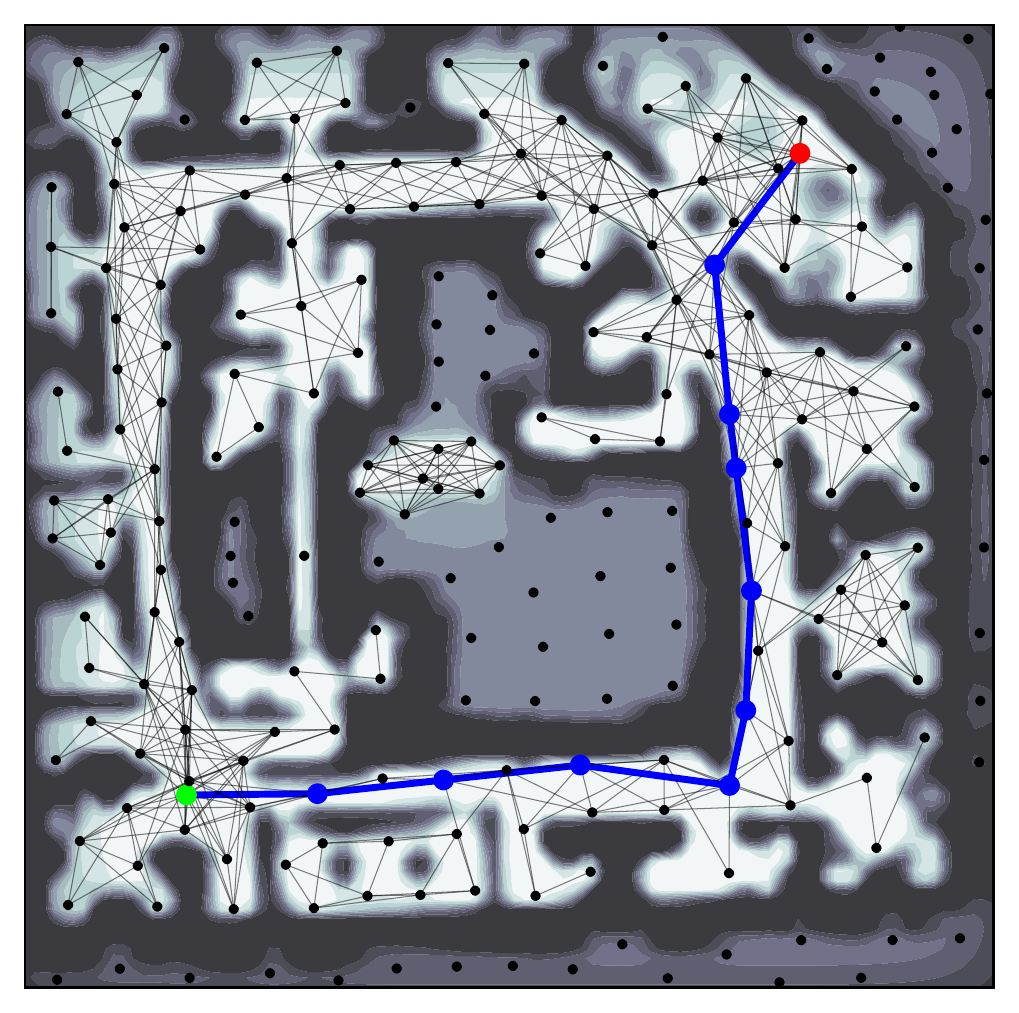}
    \caption{Navigation through \intel lab}
    \label{subfig:main_bhm}
  \end{subfigure}
  \hfill
  \begin{subfigure}[b]{0.47\linewidth}
    \centering
    \includegraphics[height=10\baselineskip]{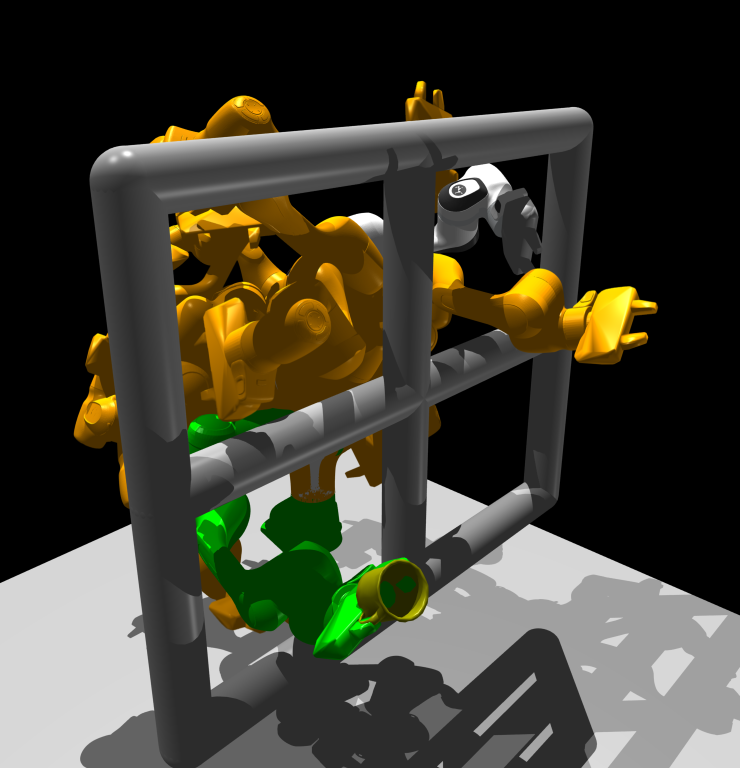}
    \caption{Cubby Reaching Manipulation Task}
    \label{subfig:main_franka}
  \end{subfigure}
\caption{We generate PRMs by sampling from a posterior feasibility distribution using particle-based Variational Inference. Placement of graph vertices is governed by a gradient flow, promoting uncertainty-guided exploration of the state space prior to graph construction. The approach results in sample-efficient path planning using both \subref{subfig:main_bhm} real-world probabilistic occupancy maps and \subref{subfig:main_franka} cost functions for manipulation tasks. In \cref{subfig:main_franka}, particle vertices are represented by orange robot configurations, with the goal configuration shown in green.}\vspace{-15pt}
\label{fig:main}
\end{figure}

Thus, roadmaps should be sparse while preserving the connectivity of the underlying free configuration space.
Our key insight is that
the distribution of roadmap samples should match the distribution over feasible states.
We formulate this problem as particle-based variational inference~\cite{liu2016svgd}.
Our Stein Variational Probabilistic Roadmap (SV-PRM) approach
deterministically optimizes the samples of a sparse roadmap
to efficiently cover the space.
This method only requires that the distribution is differentiable,
an assumption satisfied by 
continuous Bayesian occupancy maps~\cite{ocallaghan2012gpom,ramos2016hilbert,senanayake2017bhm}
from simultaneous localization and mapping (SLAM)
or otherwise assumed by
popular trajectory optimization algorithms~\cite{ratliff2009chomp,zucker2013chomp,mukadam2016gpmp,mukadam2018gpmp}.
We demonstrate the efficacy of our approach on 2D navigation and 7D manipulation problems (\cref{fig:main}).

A complementary thread of work biases the PRM sampling distribution to focus on important regions of the configuration space.
These approaches typically require additional uniform samples to provide baseline coverage of easily-sampled regions.
Heuristics from collision geometry or topology are most effective on 2D problems and increase the expense of generating a single sample~\cite{boor1999gaussian,holleman2000medial,hsu2003bridge,saroya2021neuralgas}.
Recent work has trained neural networks to propose samples around predicted shortest paths~\cite{ichter2018learning,qureshi2021mpnet} or bottleneck nodes~\cite{kumar2019lego}.
However, these approaches primarily focus on the single-query setting, i.e. with a specific start and goal configuration pair.
Critical PRMs~\cite{ichter2020critical} are sparse roadmaps designed for the same multi-query setting targeted by SV-PRMs.
This approach biases toward samples with high betweenness centrality,
training a neural network to predict such nodes from local environment features.
Neural network-based approaches require a dataset of similar planning environments,
while SV-PRMs perform variational inference with a differentiable probabilistic feasibility function.

A key question with biased samplers is the choice of environment representation.
This is often in the form of an occupancy map~\cite{kumar2019lego,saroya2021neuralgas},
point cloud~\cite{qureshi2021mpnet},
or other complete representation~\cite{ichter2018learning}.
Explicitly constructing these representations in configuration space can be challenging for many robotics settings,
especially with partial-observability.
In contrast, classical geometry heuristics only require access to a binary collision checker to evaluate candidate samples~\cite{boor1999gaussian,holleman2000medial,hsu2003bridge},
while Critical PRMs use local features to predict criticality~\cite{ichter2020critical}.
Similarly, SV-PRMs rely only on access to a differentiable probabilistic feasibility function, and avoid explicitly evaluating the function over the entire state space. 

SV-PRM relies on the SVGD method for variational inference (\cref{sec:svgd}). Although SVGD has been used for model-predictive control~\cite{lambert_stein_2020, barcelos2021dual} and trajectory optimization~\cite{lambert_stein_2020,lambert2021entropy}, we leverage its favorable properties for generating global planning graphs.



\setlength{\fboxrule}{0pt}
\setlength{\fboxsep}{0pt}

\setlength{\fboxrule}{0.2pt}
\setlength{\fboxsep}{3pt}

\section{Probabilistic Roadmaps and Safety}
\label{sec:roadmaps}

In the conventional PRM algorithm~\cite{kavraki1996prm}, a graph $G=\{V, E\}$ with vertices $V$ and edges $E$ is generated by (1) randomly sampling a vertex candidate from the continuous configuration space $\config^i \sim \configspace$, (2) performing a collision check on the sample, (3) adding $\config^i$ to $V$ if $\config^i$ is collision-free, and (4) connecting $\config^i$ to existing neighboring vertices in the graph $\config^j\in N(\config^i)$ if the corresponding edge $e_{ij}: \config^i\config^j$ is also collision-free. A path on $G$ connecting start and goal vertices $(\config_s, \config_g)\in V$ is defined as a sequence of connected vertices $\xi = (\config_1, \config_2, ..., \config_f) $ where $\config_1=\config_s$, $\config_f=\config_g$. The edge cost is a function $c : E\rightarrow \real^+$, and the path cost is the sum of edge costs over the path: $c(\xi) = \sum_{e\in\xi}c(e)$. Denoting the set of paths on $G$ as $\Xi$, planning on $G$ is then the problem of finding the lowest-cost path $\xi^*=\argmin_{\xi\in \Xi} c(\xi)$. 

A vertex $\config^i \in \mathcal{X}$ is deemed \textit{feasible} (\ie~"collision-free") if it satisfies some constraint $h(\config^i)\leq 0$. The safe set is defined by all such points, $\safeset\subseteq\configspace$, and a safe path denoted by the continuous function $\xi_\safeset=\xi_\safeset(t)\in\safeset$ where $t\in\left[0, T\right]$, $\xi_\safeset(0)=\config_s$ and $\xi_\safeset(T)=\config_g$. The path cost is then the integral of point-wise costs along the path: $c(\xi_\safeset)=\int_{t=0}^T c(\xi(t)) dt$. Any start-goal pair $(\config_s,\config_g)\in \safeset$ is deemed reachable if a corresponding path $\xi_\safeset$ exists. The graph $G$ approximates the topology of $\safeset$, since it is constructed such that $V, E \in\safeset$. A "good" graph $G$ is then one which satisfies the following properties: (1) Any point $\config^i \in \safeset$ can be connected to $G$ with a set of collision-free edges $E^i=\{e_{ij}:\,\config^i\,\config^j\,\big|\,\config^j\in N(\config^i)\}$. (2) For any start and goal $\config_s, \config_g \in \safeset$, the augmented graph $G'=(V\cup\{\config_s,\config_g\}, E\cup\{E^s, E^g\})$ contains an optimal path $\xi^*$ with a cost bounding the optimal path cost in the safe set: $c(\xi^*) - c(\xi_\safeset^*) \leq \delta$,  where $\delta \geq 0 $.
For query-time efficiency, the graph $G$ should have minimal size while satisfying (1) and (2).
The sampling strategy for generating $V$ is then vital in achieving these properties.
\section{Posterior-Guided Roadmap Generation}
\label{sec:problem}

In order to generate candidate vertices in an informed and sample-efficient manner, we desire a sampling distribution having high probability in the safe set $\safeset$, and low probability elsewhere.  Specifically, we can represent the probability of a given point $\config \in \real^d$ being collision-free by the \textit{feasibility} likelihood\footnote{We use this notation for convenience, maintaining the following equivalence: $\probc{\lvar=1}{\state;\,\map} = \probc{h(\state) \leq 0}{\state;\,\map} = \probc{\state \in \safeset}{\state;\,\map}$.} $\probc{\lvar=1}{\state;\,\map}$ with parameter $\map \in \real^m$. Here, the occupancy indicator variable $z \in \{0, 1\}$ labels a given location as being in-collision ($\lvar=0$) or collision-free ($\lvar=1$). Using Bayes' Rule, we can obtain a posterior probability over collision-free space:
\begin{align}\label{eq:posterior}
\probc{\state}{z=1;\,\map} = \eta\,\probc{z=1}{\state;\,\map}\ \prob{\state}
\end{align}
where $\prob{\state}$ is a prior probability, and $\eta$ a normalizing factor. This formulation bears particular significance in the partially-observable setting, where we do not have access to ground truth occupancy labels $\lvar$, but can make measurements using a probabilistic model accounting for sensor noise. In this case, the likelihood distribution $\probc{z=1}{\state;\,\map}$ can be interpreted as a probabilistic map of unoccupied regions~\cite{thrun2005probabilistic}. However, as we shall see, such feasibility likelihoods may also correspond to negatively-exponentiated costs found within the context of  trajectory-optimization. In both cases, we can derive a target posterior distribution to inform a sampling scheme for PRMs.

\subsection{Bayesian Occupancy Maps}
\label{sec:bom}
Bayesian inference provides a useful framework for integrating observations and incorporating uncertainty to construct occupancy maps. In this case, the target space corresponds to 2- or 3-dimensional Cartesian space, \ie~$\state \in \real^2$ or $\state \in \real^3$ .
The map parameter $\map$ can be obtained by performing Bayesian inference using collected occupancy data. Given a set of state-measurement pairs $D = \{\state_m, \lvar_m\}_{m=1}^M = (\configs, \lvarv)$, a likelihood model $\probc{z=1}{\state, \weights}$ and prior $\prob{\weights}$, the posterior over model parameters can be expressed as
\vspace{-7pt}
\begin{align}\label{eq:map_posterior}
\textstyle
    \probc{\weights}{\lvarv, \configs} &= \eta\,\prod_{m=1}^M\probc{\lvar_m}{\state_m, \weights}\prob{\weights}
\end{align}
 where $\eta$ is a normalizing constant. The feasibility likelihood $\probc{z=1}{\state;\,\map}$ can then be modeled for any point in the domain $\config\in\mathcal{X}$, where the parameter $\map$ can be chosen as a  sufficient statistic from the fitted posterior in \cref{eq:map_posterior}, \eg\ $\map \equiv \expect{\probc{\weights}{\lvarv, \configs} }{\weights}=\mu_\weights$. Approximating the posterior is non-trivial, however a variety of scalable approaches exist. Bayesian Hilbert Maps~\cite{senanayake2017bhm}, for instance, model the likelihood of free-space using the sigmoid:
\begin{align}
\textstyle
    \probc{z=1}{\state, \weights} &= \left(1+\exp(\weights^\top\feats(\state)\right)^{-1}
\end{align}
with a feature vector $\feats(\state)$ of radial-basis functions.


\subsection{Feasibility Distributions in Motion Planning}
\label{sec:traj_opt}

Defining a state trajectory as the continuous-time function $\traj \triangleq \x(t):t \rightarrow \real^d$, and a start state $\x_0$, trajectory optimization aims to find the optimal trajectory $\traj^*$ which minimizes an objective functional $\mathcal{F}(\traj; \x_0)$. The solution must be feasible and avoid collisions with obstacles (\ie~belong to the safe set). This condition can be imposed by including an additional obstacle penalty $\mathcal{F}_{obs}$ in the objective~\cite{ratliff2009chomp,zucker2013chomp}:
\begin{align}
    \mathcal{F}(\traj; \x_0) = \mathcal{F}_{task}(\traj; \x_0) +  \alpha \mathcal{F}_{obs}(\traj) 
\end{align}\vspace{-5pt}
where $\mathcal{F}_{task}(\traj; \x_0)$ includes all other task objectives and constraints, such as distance-to-goal and penalty on joint-limit violations. The objective is minimized via a Gauss-Newton form of iterative gradient descent, leveraging both gradient and local curvature of the cost function for rapid convergence~\cite{ratliff2009chomp, schulman2014trajopt, mukadam2018gpmp}.
The obstacle cost at a given trajectory state $\config \in \traj$ can be modelled using a truncated signed-distance field (t-SDF), which penalizes a state if a Cartesian point on the robot is within an $\epsilon$-distance from the surface of the neareast obstacle. Approximating the robot surface using a collection of body spheres $s_j, j=1:K$, and letting $d(\config, s_j)$ to be the distance from the surface of $s_j$ to the nearest obstacle, a t-SDF value is defined by the hinge loss: \vspace{-5pt}
\begin{equation}
\textstyle
c(\config, s_j) = \begin{cases}
-d(\config, s_j) + \epsilon & d(\config, s_j) \leq \epsilon \\
0 & \mathrm{otherwise}
\end{cases}
\end{equation}
These costs are combined across spheres to construct an obstacle-cost vector: $\mathbf{h}(\config) = \left[ c(\config, s_j)\right]\big|_{j=1:K}$, and define the total obstacle cost as the scaled inner-product: $\mathcal{F}_{obs}(\config) = \alpha \norm{\mathbf{h}(\config)}^2$, where $\alpha > 0$.
As in \cite{mukadam2016gpmp,mukadam2018gpmp}, the obstacle cost at a given state can be interpreted as the negative log-likelihood function: $-\log\probc{\lvar=1}{\config;\,\map} = \alpha \norm{\mathbf{h}(\config)}^2 + \mathrm{const}$, with parameter $\map$ including obstacle locations and geometry. Here, the likelihood probability of a state being collision-free is modeled as $\probc{\lvar=1}{\config;\,\map} \propto \exp(-\alpha \norm{\mathbf{h}(\config)}^2)$. Motion planning can then be framed as an inference problem, where the objective is to find the \textit{maximum a-posteriori} solution: 
\begin{align}
    \tau^* &=\argmax_\tau \probc{\tau}{\lvar=1; \map} \label{eq:posterior_trajopt}\\
    &= \argmax_\tau \prod_{t=0}^T \probc{\lvar=1}{\config_t; \map}\  p(\tau)
\end{align}
 where trajectories are discretized into a set of states $\tau = \{\config_t\}_{t=0:T}$ over a horizon $T$.

Although gradient-based controllers (ex. iLQR) and motion planners (ex. GPMP) can quickly return a feasible trajectory, the solution may only be \textit{locally} optimal, and requires an initial trajectory or set of waypoints to prevent falling into poor optima. Such a path can be produced by sampling PRM vertices from the feasibility posterior over state-space, $\config^i \sim \probc{\config}{\lvar=1; \map} =  \eta\ \probc{\lvar=1}{\config; \map}\ \prob{\config}$, and generating the shortest-path on the resulting graph. 

For both cases in \cref{sec:bom} and \cref{sec:traj_opt}, we would  like to directly sample from the posterior distributions in order to generate viable points in regions of the state space that are likely to satisfy state-based constraints: $\config^i \sim \probc{\config}{\lvar=1; \map}$. Unfortunately, obtaining a closed-form expression for this target distribution is intractable in general. The posterior-sampling procedure must then be approximated using methods such as Markov-Chain Monte Carlo (MCMC) or Variational Inference (VI). 

\begin{figure*}[!ht]
\newlength{\bhmenvwidth}
\setlength{\bhmenvwidth}{0.24\linewidth}
\centering
\begin{subfigure}{\bhmenvwidth}
\includegraphics[trim={8 8 8 8},clip,width=\linewidth]{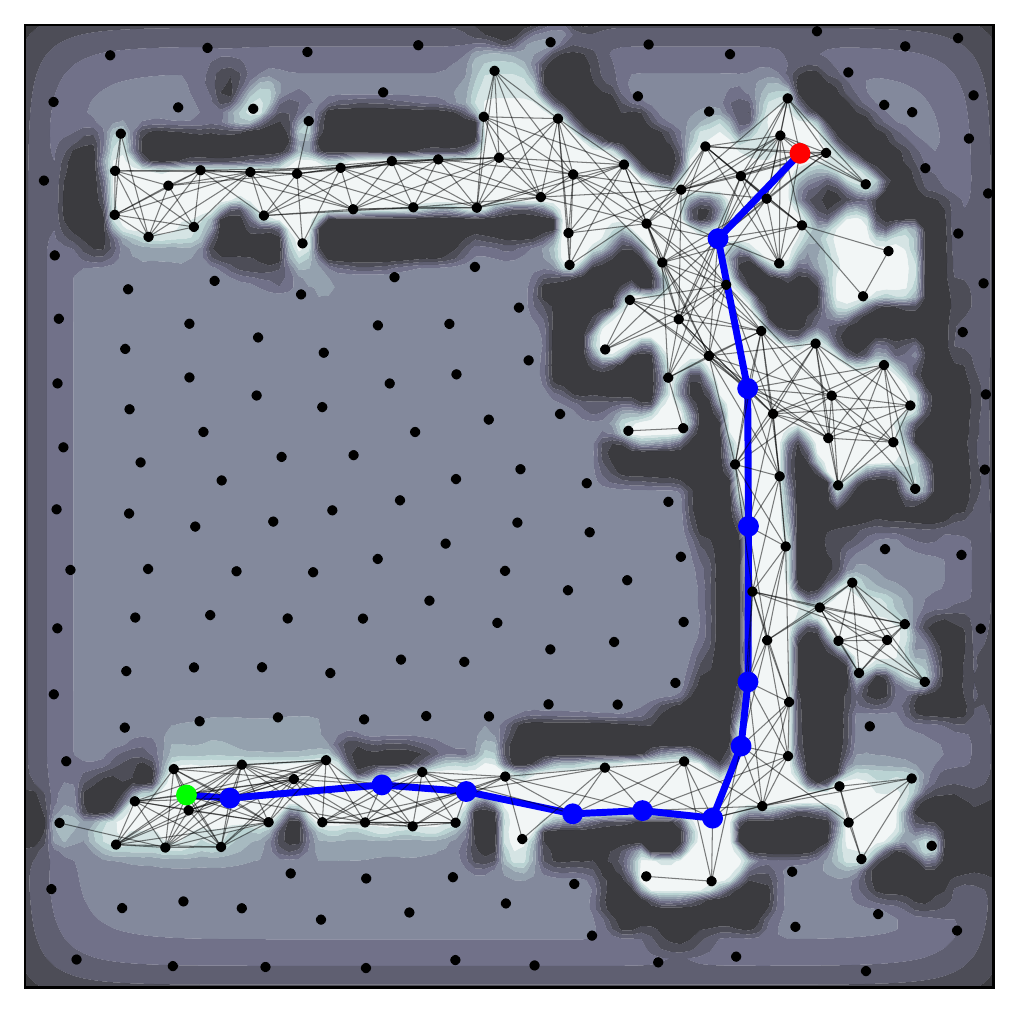}
  \caption{SV-PRM(Uniform), conservative $\thresh$.}
  \label{fig:bhm050ect25}
\end{subfigure}
\begin{subfigure}{\bhmenvwidth}
\includegraphics[trim={8 8 8 8},clip,width=\linewidth]{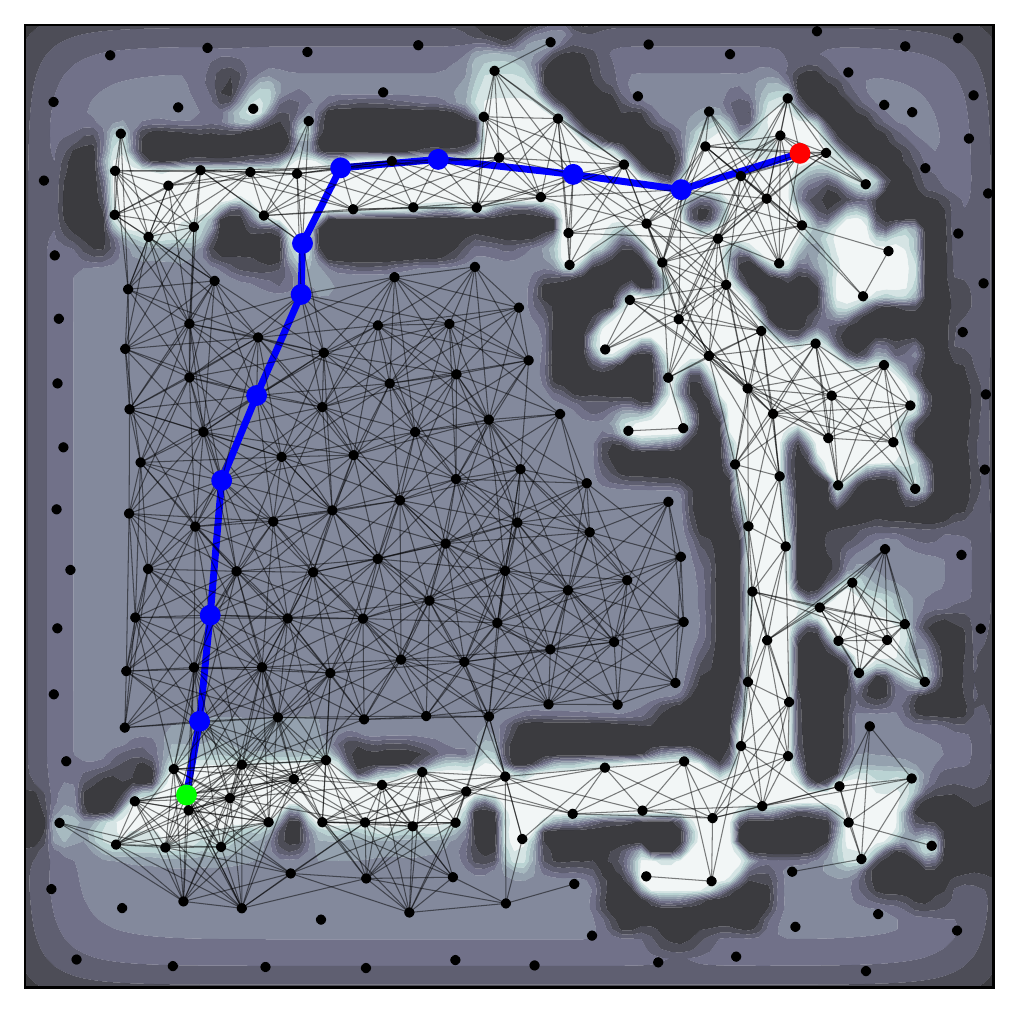}
  \caption{SV-PRM(Uniform), optimistic $\thresh$.}
  \label{fig:bhm050ect40}
\end{subfigure}
\hfill
\begin{subfigure}{\bhmenvwidth}
\includegraphics[trim={8 8 8 8},clip,width=\linewidth]{figs/bhms/900iter_cr=5_ct=5.pdf}
  \caption{SV-PRM(Uniform), conservative $\thresh$.}
  \label{fig:bhm900ect25}
\end{subfigure}
\begin{subfigure}{\bhmenvwidth}
\includegraphics[trim={8 8 8 8},clip,width=\linewidth]{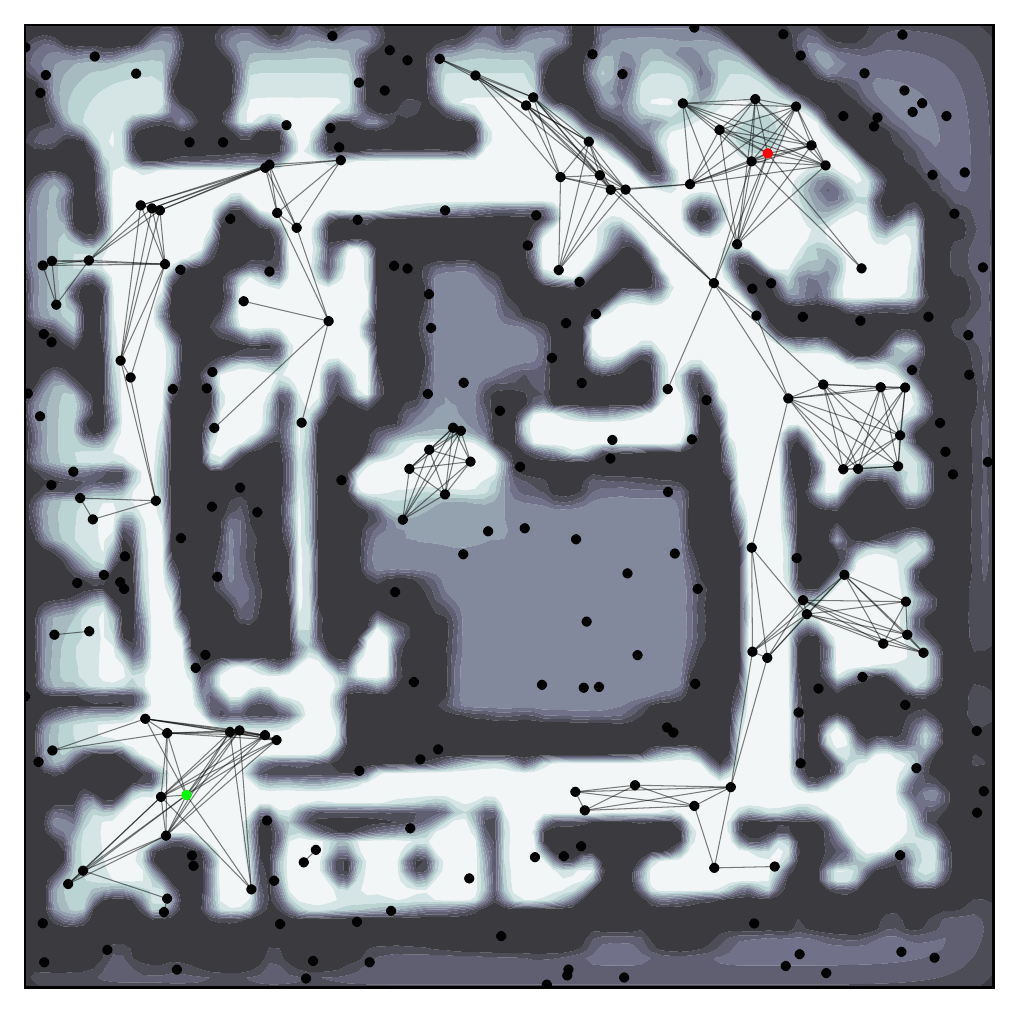}
  \caption{PRM(Uniform), conservative $\thresh$.}
  \label{fig:bhm900unif}
\end{subfigure}

\caption{
(a)-(b) Partially and (c)-(d) fully-explored \intel Bayesian Hilbert Map environment. (a)-(b) compares the resulting SV-PRM with conservative $\thresh = 0.54$ and optimistic $\thresh = 0.45$ graph generation. (c) demonstrates the improved connectivity of SV-PRM(Uniform) relative to (d) the PRM(Uniform) particles (which were used to initialize the SV-PRM in (c). }
\label{fig:bhms_partial}
\vspace{-1\baselineskip}\vspace{-5pt}
\end{figure*}

\section{Variational Inference and SVGD}
\label{sec:svgd}

Variational inference (VI) is a powerful tool for approximating challenging probability densities in Bayesian statistics.  As opposed to MCMC methods, VI formulates inference as an optimization problem. A proposal distribution $q(\state)$, belonging to a family $\mathcal{Q}$, is chosen to minimize the KL-divergence with the target posterior distribution $p(\state\,|\lvar)$ over latent variable $\state$: 
\begin{align}
\textstyle
 q^*(\state) &= \min_{q\in\mathcal{Q}} \dkldiv{q(\state)}{p(\state\,|\,\lvar)}
\end{align}

Traditional VI methods typically require careful selection of the distribution class $\mathcal{Q}$, which is often chosen to have a tractable parameteric form at the expense of introducing bias. 
A recently developed method,\textit{ Stein Variational Gradient Descent}~\cite{liu2016svgd,wang2019matrixsvgd}, avoids the challenge of determining an appropriate $\mathcal{Q}$ by leveraging a non-parameteric, particle based representation of the posterior distribution.  This approach approximates a posterior $\probc{\state}{\lvar}$  with a set of particles $\{\state^i\}_{i=1}^{N}$, $\state^i \in \real^d$. The particles are iteratively updated according to $\state^i \leftarrow \state^i + \epsilon \bm{\phi}^*(\state^i)$,  given a step-size $\epsilon$. 
The function $\bm{\phi}^*(\cdot)$ lies in the unit-ball of a reproducing kernel Hilbert space (RKHS). This RKHS is characterized by a positive-definite kernel $k(\cdot, \cdot)$. The term $\bm{\phi}^*(\cdot)$ represents the optimal perturbation or velocity field (i.e. gradient direction) which maximally decreases the KL-divergence:
\begin{align}
\bm{\phi}^*  = \argmax_{\bm{\phi} \in \mathcal{H}} \Big\{ - \nabla_\epsilon \dkldiv{ q_{\left[\epsilon \bm{\phi}\right]}}{p(\state\,|\,\lvar)}\,\mathrm{s.t.}\, \norm{\bm{\phi}}_{\mathcal{H}} \leq 1\Big\},
\end{align}
where $q_{\left[\epsilon \bm{\phi} \right]}$  indicates the particle distribution resulting from taking an update step. This has been shown to yield a closed-form solution~\cite{liu2016svgd}
which can be interpreted as a functional gradient in RKHS, and can be approximated with the set of particles: \vspace{-5pt}
\begin{align}\label{eq:phi_hat}
\hat{\bm{\phi}}^*(\state) = \frac{1}{N}\sum_{j=1}^{N}
\Big[
k(\state^j, \state)\nabla_{\state^j}\log \probc{\state^j}{\lvar}
+ \nabla_{\state^j} k(\state^j, \state) \Big].
\end{align}
\cref{eq:phi_hat} has two terms that control different aspects of the algorithm. The first term is essentially a scaled gradient of the log-likelihood over the posterior's particle approximation. The second term is known as the {\em repulsive force}. Intuitively, it pushes particles apart when they get too close to each other and prevents them from collapsing into a single mode. This allows the method to approximate complex, possibly multi-modal posteriors. For the case of a  single particle, the method reduces to a standard optimization of the log-likelihood or a MAP estimate of the posterior as the repulsive force term vanishes, \textit{i.e.} $\nabla_{\state} k(\state, \state) = 0$. SVGD's optimization structure empirically provides better particle efficiency than other popular sampling procedures, such as Markov Chain Monte Carlo~\cite{chen2019steinpm}. The deterministic, gradient-based updates result in smooth transformations of the proposal distribution, a property which makes SVGD particularly attractive for trajectory optimization and inference.

\subsection{Hessian-Scaled Kernels}
\label{sec:hess_svgd}

As discussed in \cite{detommaso2018svn, wang2019matrixsvgd}, the convergence and accuracy of the SVGD algorithm can be largely improved by incorporating curvature information into  the kernel. For instance, a positive-definite matrix $M$ can be used as a metric to scale inter-particle distances inside of an \textit{anisotropic} RBF kernel: $k(\state^j, \state^i) = \exp \big(-\frac{1}{2h}(\state^j - \state^i)^\top M (\state^j - \state^i) \big)$, where $h$ is the bandwidth parameter. Curvature information can then be shared across particles by averaging their local Hessian evaluations. Specifically, denoting the negative Hessian matrix to be $H(\state) = -\nabla^2_\state \log p(\state\,|\,\lvar)$, we can define the metric $M = \frac{1}{N}\sum_{j=1}^{N} H(\state^j)$, which is computed using $\state^j$-values from the previous iteration. 

\section{Stein Variational PRMs}
\label{sec:method}

As an alternative to random or purely heuristic-driven sampling for PRMs, we propose to approximate the posterior distribution of feasible space $\probc{\state}{\lvar=1;\,\map}$ using a set of SVGD particles  $\{\state^i\}_{i=1}^{N}$. 

In the \textit{Inference} phase, particles are initialized by drawing samples from the prior distribution over configuration space: $\{\config^i\}_{i=1}^{N} \sim \prob{\config}$. The particle distribution is then updated using successive SVGD iterations, where at each iteration, the log-likelihood and log-prior gradients are computed for each particle \textit{in parallel}:  
\begin{align}\label{eq:svprp_grad}
 \nabla_\state \log \probc{\state^i}{&\lvar=1;\,\map} =\\
 &\nabla_\state \log \probc{\lvar=1}{\state^i;\,\map} + \nabla_\state \log \prob{\state^i} ,
\end{align}
Given a choice of kernel $k(\cdot, \cdot)$, the RKHS gradient in \cref{eq:phi_hat} is computed for each particle. The particles are then updated, and the process is repeated until convergence.

During the \textit{Construction} phase, the feasibility of a given point $\config$ is imposed by a chance constraint on the feasiblity-likelihood: $\probc{\lvar=1}{\config;\,\map} \geq \thresh$, where $\thresh\in[0, 1]$. Vertices and edges are only accepted into the graph $G$ if this constraint is satisfied, ensuring that they lie within the safe set $\safeset$ with a probability of at least $\thresh$. Particles which violate the chance-constraint are removed from the vertex-candidate set. Edges are connected between two vertices if the edge-length is less than the connectivity radius $\conrad$, which defines the set of nearest neighbors for each vertex. In practice, this can drastically reduce the number of edges in the graph, therefore decreasing the problem size in the planning phase.

In the \textit{Planning} phase, a feasible path is generated by first performing a "collision-check" on a candidate edge to ensure that the constraint $\probc{\lvar=1}{\config;\,\map} \geq \thresh$ holds.  Equi-distant  query points are generated along the edge at a fixed resolution, and edge-feasibility is approximated by ensuring that all query points satisfy the chance constraint. If the edge is accepted in to the graph $G$, its length is evaluated and stored. A user-specified search algorithm is then run to completion in order to generate a shortest-path solution $\xi^*$.
 

A subtle but key aspect of this VI-based PRM approach is that a candidate set of vertices are \textit{not} generated by first fitting a parameterized distribution $q(\config)$ to the target posterior $\probc{\config}{\lvar=1;\,\map}$, then sampling points from $q(\config)$. Rather, the particles themselves both govern the implicit empirical distribution $q$ and comprise the set of candidate vertices. As these vertices are \textit{deterministically} optimized according to the SVGD algorithm, they will approach regions of the space which are more likely to be collision-free, however the repulsive-force term will ensure particle diversity and adequate coverage of the target distribution. The non-parametric nature of the algorithm allows this approach to be flexible and model highly multi-modal distributions.

This uncertainty-guided procedure for PRM generation is quite general, and does not necessitate sampling heuristics. However, the prior distribution $\prob{\config}$ can be chosen to bias portions of the configuration space if required. Furthermore, this approach can exploit gradient-based information of the target distribution, and can be implemented in batch for efficient GPU parallelization. 

With the $\thresh$-parameter, we can control optimism in graph construction. Low values can lead to graphs which populate edges in uncertain or unexplored areas, producing optimistic path solutions with shorter paths. Conversely, higher values restrict vertices and edges to high-probability regions, resulting in longer paths. This behavior is depicted in \cref{fig:bhms_partial} with a partial occupancy map generated from the Intel dataset.

Although SV-PRM requires a differentiable probability distribution to evaluate the gradient in \cref{eq:svprp_grad}, as discussed in \cref{sec:problem}, this property can be satisfied with continuous Bayesian occupancy maps or is already assumed by popular trajectory optimization algorithms. 

\section{Experiments}
\label{sec:experiments}

We evaluate Stein Variational PRMs and relevant baselines on three planning environments:
(1) a synthetic planar point-navigation problem with challenging obstacle distributions,
(2) a probabilistic occupancy map generated from a real-world baseline,
and (3) a high-dimensional simulated manipulation problem.  
In all experiments, we use the anisotropic RBF kernel (\cref{sec:svgd}). For 2D examples with uniform priors, we use the following positive-definite metric to scale the pairwise particle distances: 
$M = \frac{1}{N}\sum_{i=1}^{N} \nabla_\state\log p(\lvar=1\,|\,\state^i)\nabla_\state \log p(\lvar=1\,|\,\state^i)^\top$.

We evaluate different roadmap generation strategies,
where the traditional PRM generates samples according to a (pseudo-)random number generator,
potentially with cost-based rejection sampling.
SV-PRM is initialized with the samples from PRM,
then performs SVGD on the particle set.
The connection strategy described in \cref{sec:method} is used for both roadmaps.
The \lazysp~\cite{dellin2016lazysp} algorithm is then used to search the roadmap,
lazily evaluating edges to check whether the cost threshold would be exceeded.
We parallelize SVGD evaluations using a PyTorch implementation, allowing efficient GPU-driven inference.  
\subsection{Navigation}

In the synthetic \checkerboard environment (\Tabref{tab:toy2dqual}),
we compare performance in low- and high-particle regimes across various initialization schemes.
The continuous target distribution is generated by fitting a grid of RBF features to a binary occupancy map.
In the real world \intel environment (\cref{subfig:main_bhm}), we fit a Bayesian Hilbert Map (BHM)~\cite{senanayake2017bhm} to LIDAR data from the Intel Research Lab in Seattle~\cite{Radish}.
The resulting probabilistic occupancy map is the SV-PRM target distribution.
We introduce a barrier function to prevent particles from exceeding the limits of configuration space. \slnote{write equations.}


\subsection{7-Dof Motion Planning}

To demonstrate the scalability of SV-PRMs in higher-dimensional robotics problems,
we test the approach on a simulated reaching task depicted in \cref{subfig:main_franka}.
Here, the start pose of the modeled Franka arm is configured to be reaching in the top-right compartment.
A feasible path in joint-space $\config\in\real^7$ must be found to reach the goal configuration (green) in the lower left compartment, while avoiding the cabinet frame (grey).
Similarly to \cref{sec:traj_opt}, the obstacle cost $\mathbf{h}(\config)$ is modeled using a smoothly varying t-SDF, as used in \cite{zucker2013chomp}, with an offset value of $0.25\mathrm{m}$,
and a feasibility likelihood of $\exp(-10\norm{\mathbf{h}(\config)}^2)$.
The prior $\prob{\config}$ is formulated as a uniform distribution,
with exponentially decreasing probability near the system joint-limit values. 
As in the Navigation experiments, we compare our approach to traditional PRMs, varying the initialization type between random uniform samples and pseudo-random samples generated from a Halton sequence.
Further, we add a task-based heuristic sampling distribution using a multi-modal mixture of Gaussians.
Three isotropic Gaussian components are centered on the start and goal configurations, as well as on a retracted "home" away from the obstacles.

\section{Results}
\label{sec:results}



\begin{table}[t]
\newlength{\heatmapwidth}
\setlength{\heatmapwidth}{0.3\linewidth}
\centering
\setlength\tabcolsep{1.5pt} 
\begin{tabular}{cccc}
        &   PRM(U+Rej)& PRM(H+Rej) & SV-PRM(U+Rej) \\
\rotatebox{90}{\quad\quad\quad N=100} &
\includegraphics[trim={8 8 8 8},clip,width=\heatmapwidth]{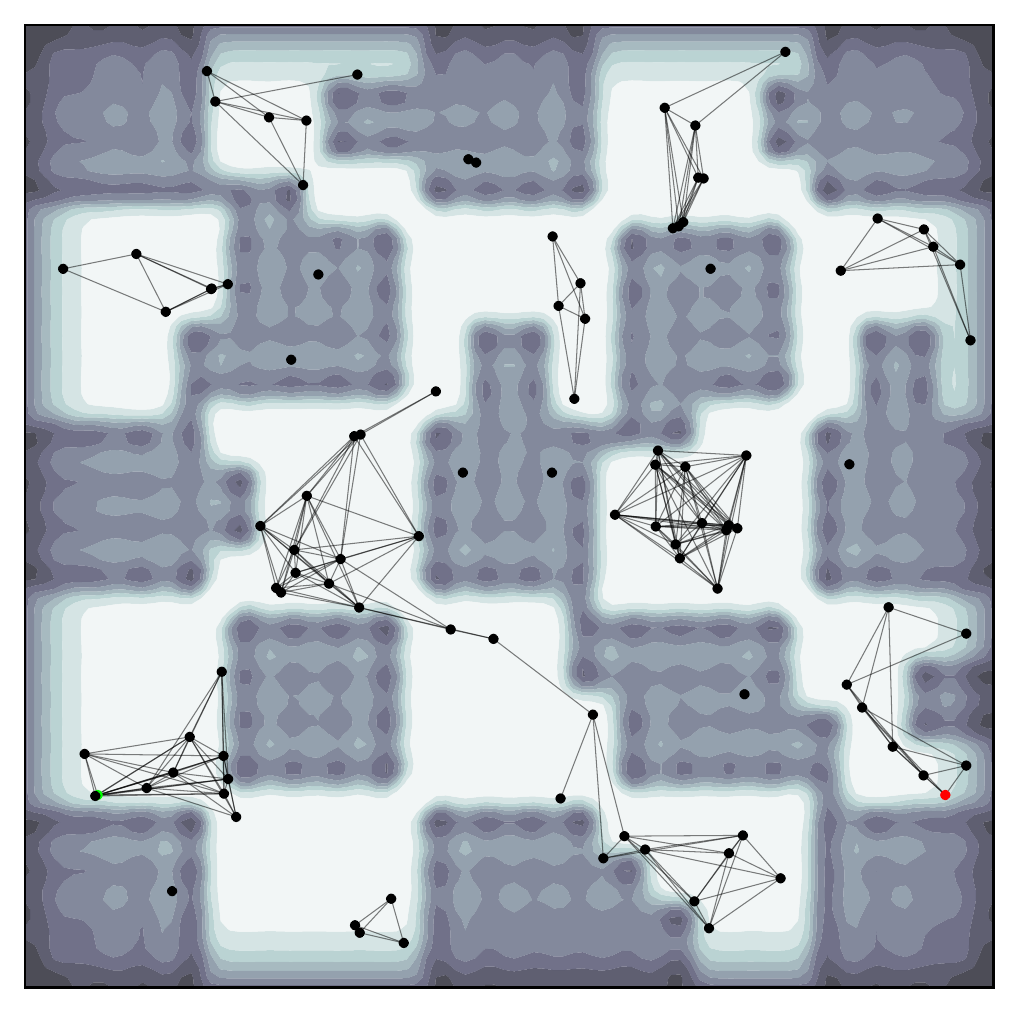} &
\includegraphics[trim={8 8 8 8},clip,width=\heatmapwidth]{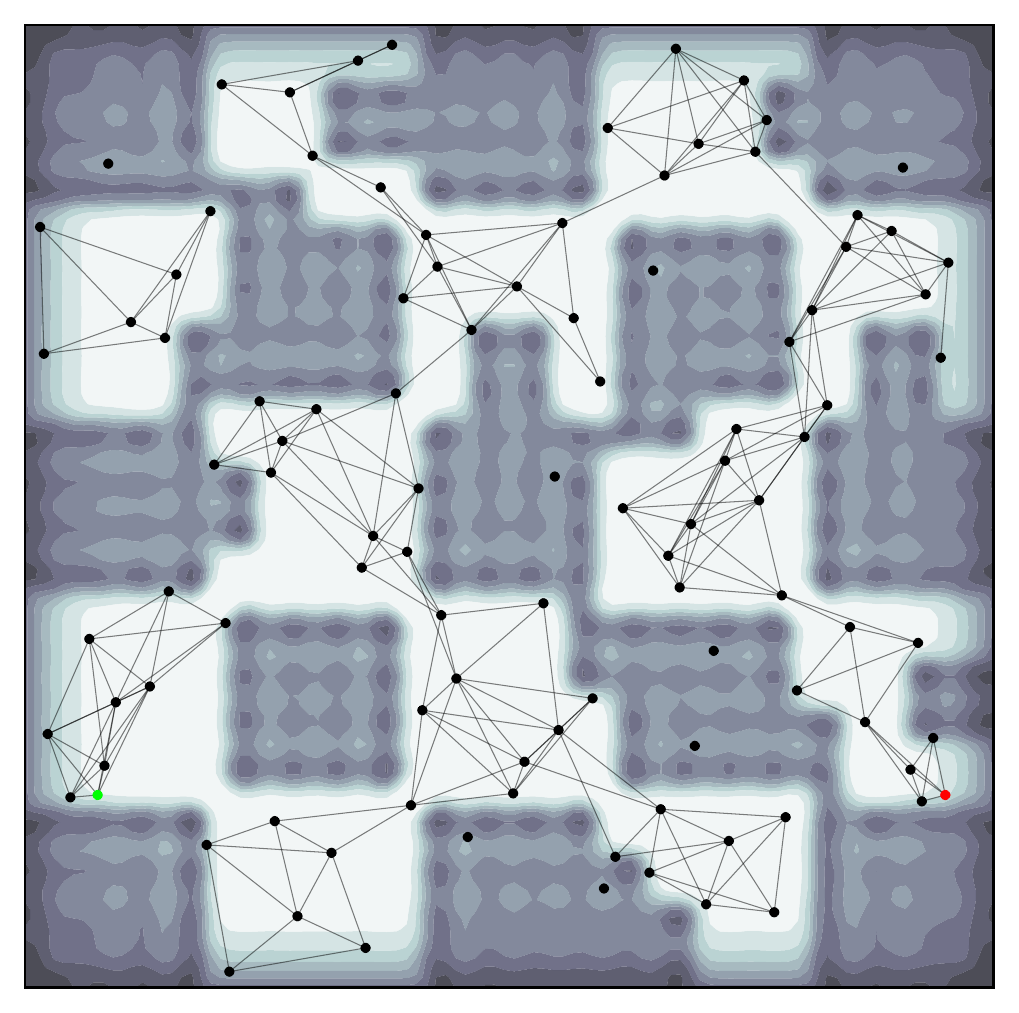} &
\includegraphics[trim={8 8 8 8},clip,width=\heatmapwidth]{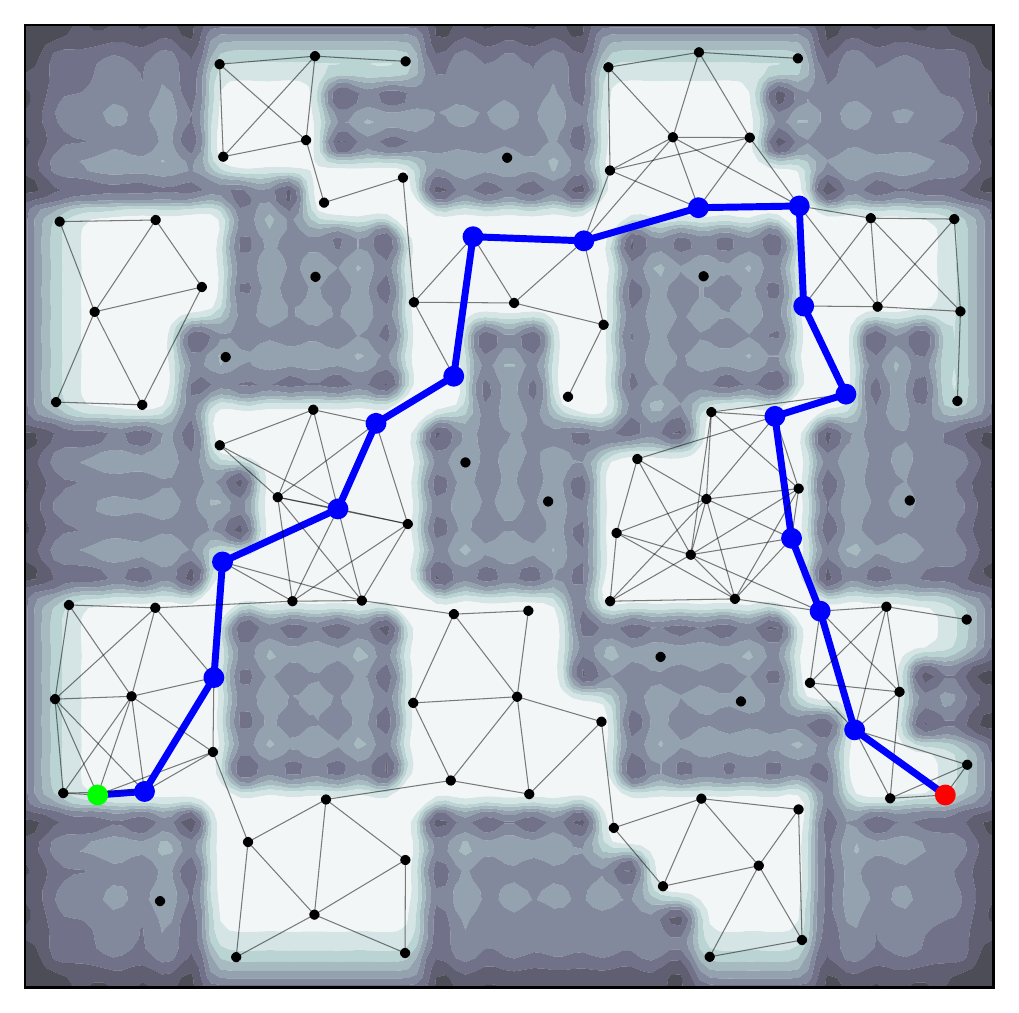} \\ 
\rotatebox{90}{\quad\quad\quad N=250} &
\includegraphics[trim={8 8 8 8},clip,width=\heatmapwidth]{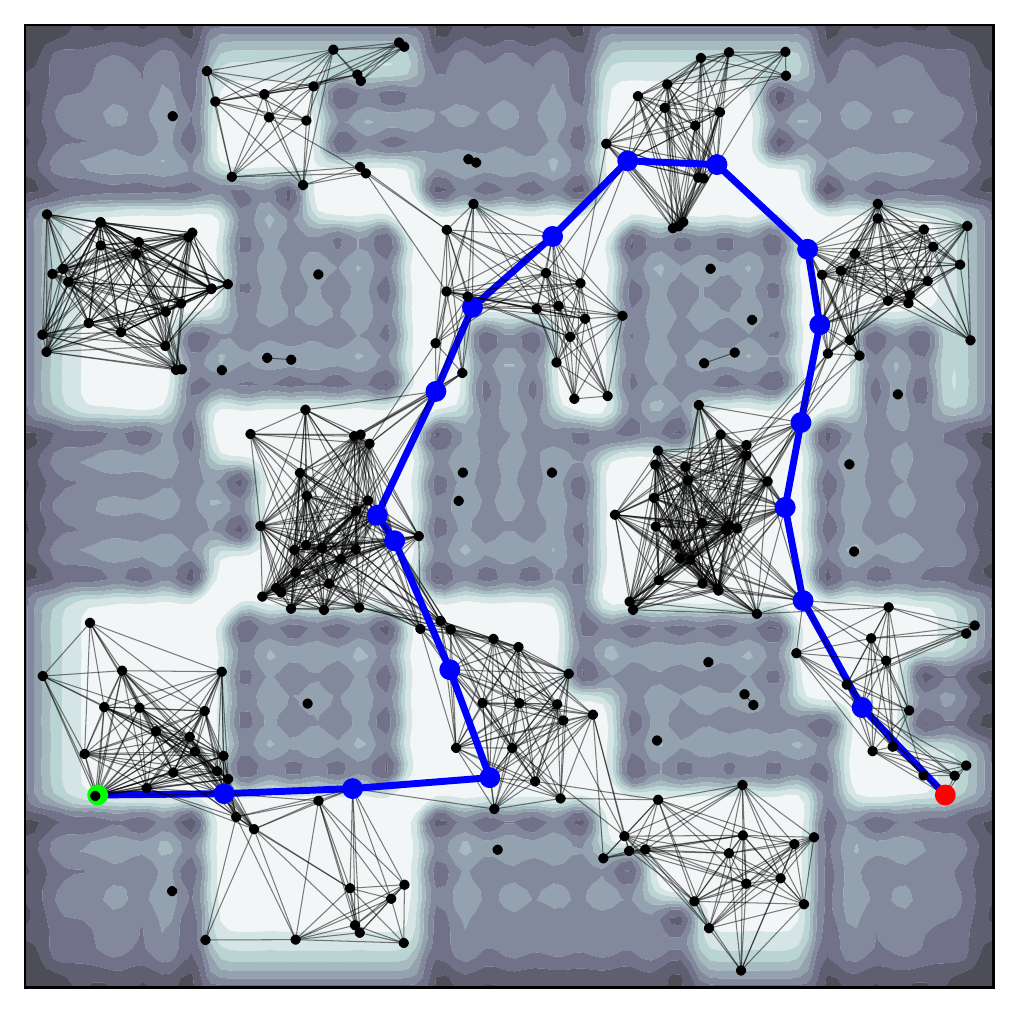} &
\includegraphics[trim={8 8 8 8},clip,width=\heatmapwidth]{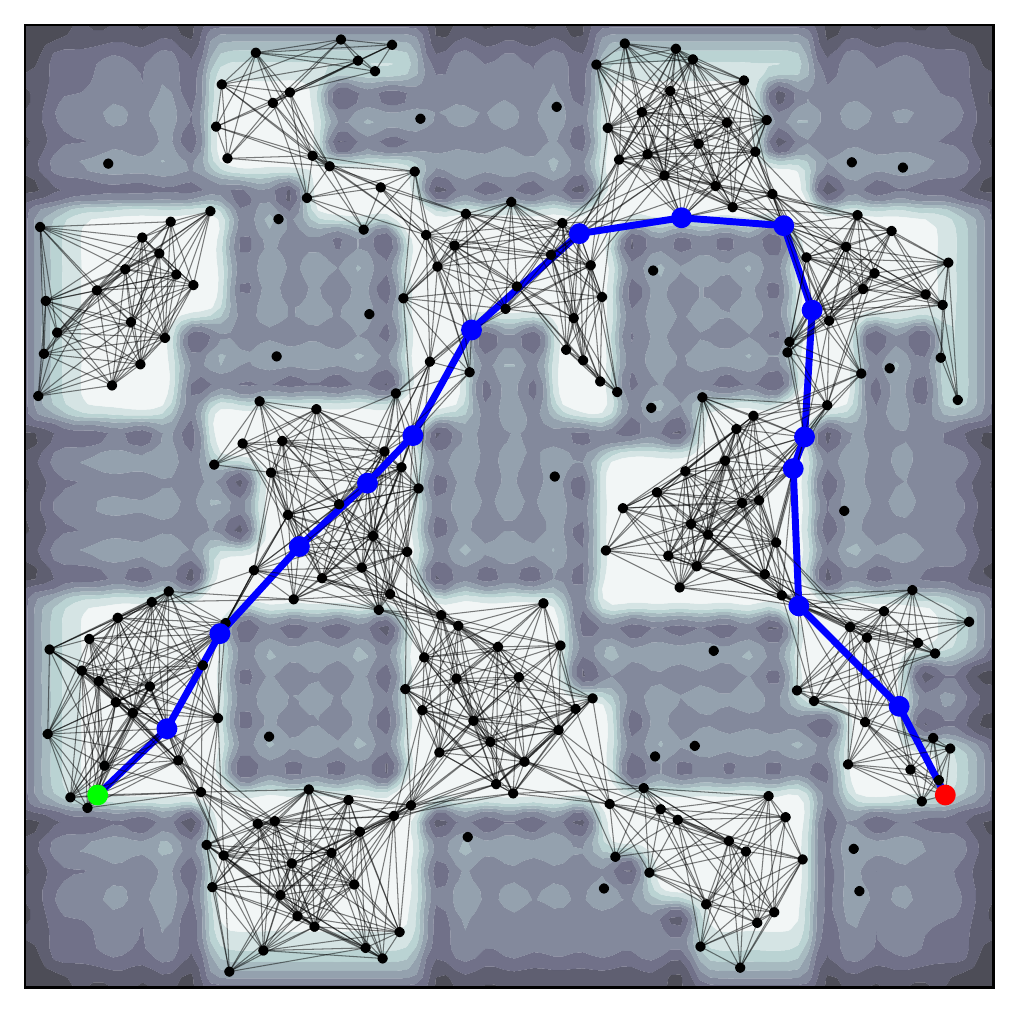} &
\includegraphics[trim={8 8 8 8},clip,width=\heatmapwidth]{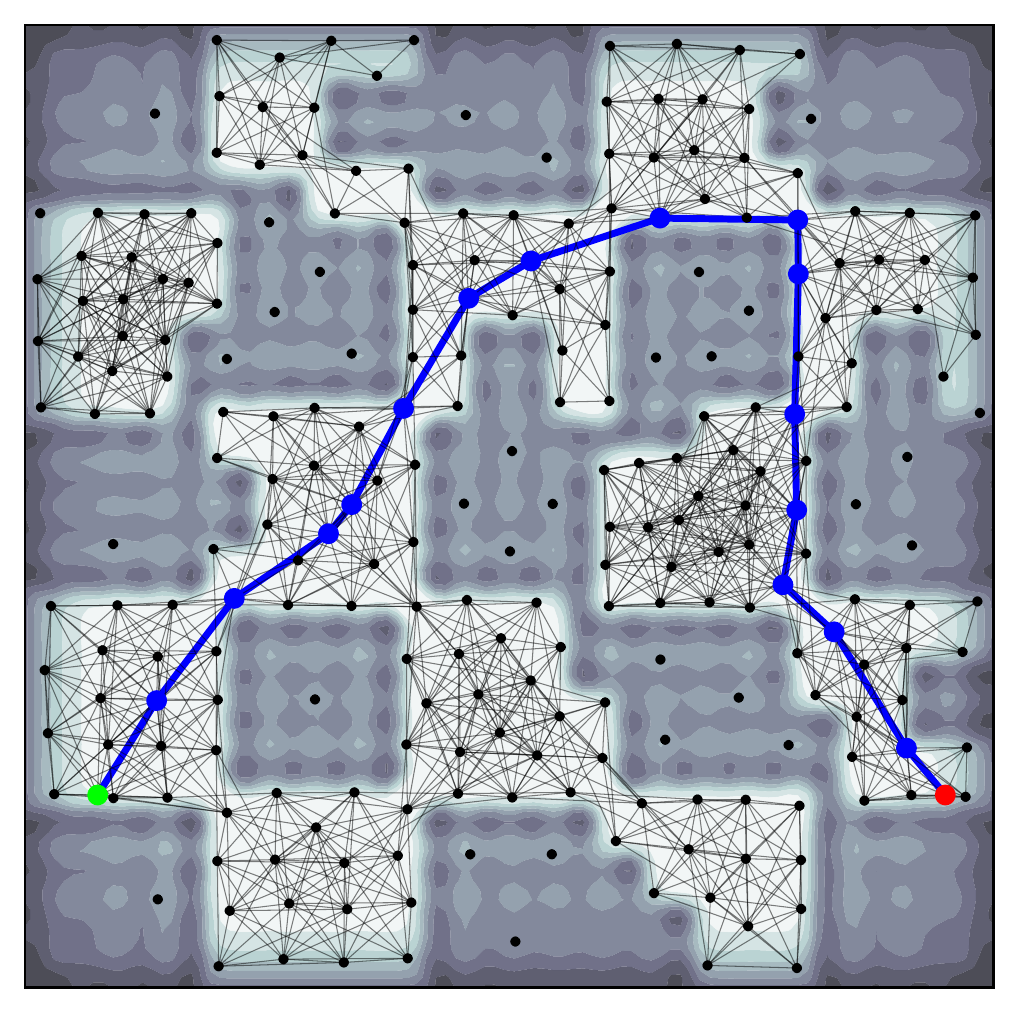} \\ 
\end{tabular}
\caption{\checkerboard environment roadmaps generated with different sampling strategies and number of samples.
The planning query consists of the start (green) and goal (red),
and the shortest path (blue) is visualized if it exists.
Best viewed digitally.
}
\label{tab:toy2dqual}
\end{table}

%

\begin{table}[t]
\centering
\renewcommand{\arraystretch}{1.1}
\begin{tabular}{@{}lrrrrc@{\hspace{0.5em}}rrrr@{}}
\toprule
\multirow{2}{*}{} & \multicolumn{4}{c}{PRM} & & \multicolumn{4}{c}{SV-PRM}\\
\cmidrule{2-5} \cmidrule{7-10}
        &  U & U+Rej & H       & H+Rej & & U  & U+Rej & H       & H+Rej \\
\midrule                                           
N = 100 &  2 & 15    & \xmark  & \cmark & & 29 & 30    & \cmark  & \cmark \\
N = 250 & 20 & 30    & \cmark  & \cmark & & 30 & 30    & \cmark  & \cmark \\
\bottomrule
\end{tabular}
\caption{
Number of successful planning trials on the \checkerboard environment with N=100 and 250 sampled vertices.
Uniform-based samplers are evaluated across 30 random seeds.
The low-discrepancy Halton-based sampler~\cite{halton1960sequence} is deterministic, so success is denoted with a \cmark or \xmark.
In the low-particle regime, uniform samples (with or without rejection) rarely capture a path between the start-goal pair.
Despite these poor initializations, the SVGD optimization yields roadmaps with feasible paths in all but one trial.
\rmsnote{TODO keep this close to \Tabref{tab:toy2dqual} at final formatting}
}\vspace{-15pt}
\label{tab:toy2d_success} 
\end{table}

We report \textit{query-dependent} metrics on the \checkerboard environment,
which measure the benefits of using the SV-PRM for specific start-goal planning queries.
We select a challenging start-goal pair that requires the roadmap to pass through several narrow passages (\Tabref{tab:toy2dqual}).
\Tabref{tab:toy2d_success} reports the number of successful planning trials for low- and high-particle regimes,
showing that SV-PRMs better capture the connectivity of the space with the same number of samples.
Compared to the larger PRMs that are needed to match the smaller SV-PRMs connectivity,
this improved sample efficiency and sparsity yields faster planning times and statistically-equivalent path lengths.
\bhnote{These full results will be available in the appendix.}

\newlength{\compressedheight}
\setlength{\compressedheight}{7\baselineskip}

\begin{figure}[!t]
\centering
\includegraphics[width=\linewidth]{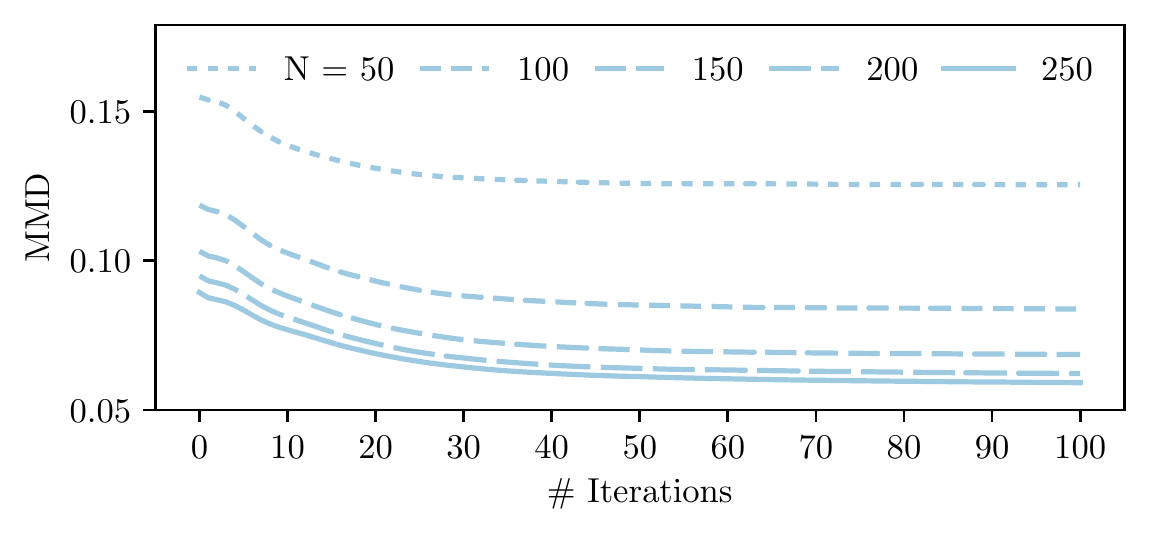}
\caption{
  Maximum mean discrepancy (MMD) of SV-PRM(Uniform) with respect to the uniform feasible distribution of the ground truth fully-mapped Intel-BHM model (\Figref{fig:bhm900ect25}).
  This differs from the partial map that SV-PRM optimizes the particles against,
  but demonstrates how particles have been distributed throughout the unmapped regions.
  Results are averaged across 100 trials.
  MMD decreases with SVGD iterations, and with increasing particle number $N$.
}
\label{fig:svprm_unif_iters_vs_mmd}
\vspace{-1\baselineskip}\vspace{-10pt}
\end{figure}

\begin{figure}[!t]
\centering
\includegraphics[width=\linewidth]{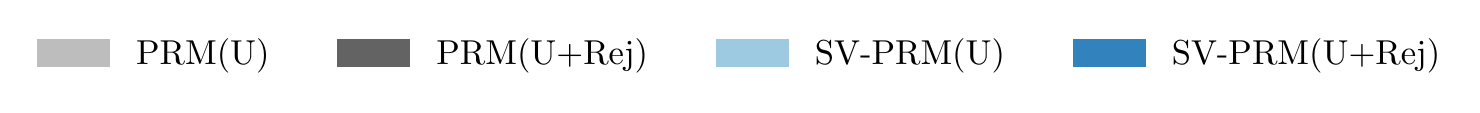}
\includegraphics[width=\linewidth]{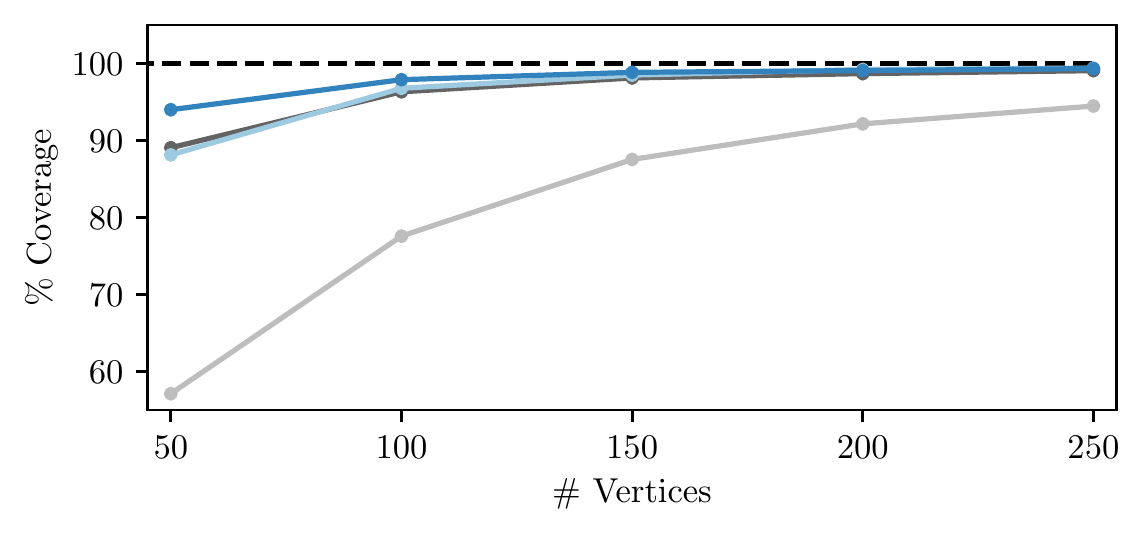}
\caption{
  Roadmap coverage as a function of roadmap size,
  estimated by sampling 1000 collision-free states in the partial Intel-BHM map
  and attempting to connect them to roadmap vertices via collision-free edges.
  SV-PRM(Uniform) and PRM(Uniform+Rejection) achieve comparable coverage,
  while SV-PRM(Uniform+Rejection) achieves slightly higher coverage with greater improvement in the low-particle regime.
}
\label{fig:coverage}
\vspace{-1\baselineskip}\vspace{-5pt}
\end{figure}

On the \intel environment,
we additionally consider \textit{query-independent} metrics that evaluate the approximation quality of the resulting roadmap
without a specific planning query context.
These metrics help understand the multi-query performance of SV-PRM.

First, we simulate a partial map by limiting the amount of data ingested by the BHM.
\Figref{fig:bhms_partial} visualizes the map when the BHM has consumed little data;
the outer hallway loop has yet to be identified.
The SV-PRM on this partial map reflects the resulting uncertainty in the model.
Many vertices are in the likely collision-free spaces,
but vertices are also distributed throughout the unmapped (and therefore uncertain) area.
By choosing the cost threshold appropriately, 
SV-PRMs preserves more vertices and edges in the roadmap.
The same SV-PRM can yield multiple candidate paths for a specific query by varying the threshold:
a conservative threshold safely connects the start and goal via the known corridor,
while an optimistic threshold navigates through the unknown and finds a shorter (but ultimately invalid) path.
In this way, users are able to specify their degree of certainty in the map quality.
SV-PRMs cull vertices accordingly to efficiently plan paths at the desired cost threshold.

To further characterize this partial map setting,
we evaluate the maximum mean discrepancy (MMD)~\cite{gretton2008mmd},
which measures the difference between two distributions based on samples drawn from each of them.
\Figref{fig:svprm_unif_iters_vs_mmd} visualizes the MMD as particles are optimized against the partial map,
demonstrating that SVGD improves the MMD over time.
Increasing the number of particles also lowers the MMD, although with diminishing returns.

\begin{figure}[!t]
\centering
\includegraphics[width=\linewidth]{bhm_plots/horizontal.pdf}

\hfill
\begin{subfigure}{0.49\linewidth}
\centering
\includegraphics[height=\compressedheight]{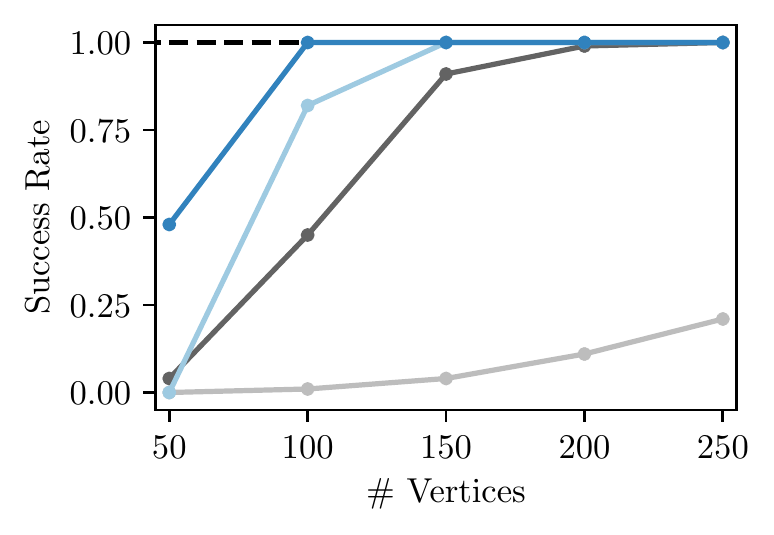}
\caption{Rate of trials with feasible solution.}
\label{fig:num_samples_vs_percent_solved}
\end{subfigure}
\hfill
\begin{subfigure}{0.49\linewidth}
\centering
\includegraphics[height=\compressedheight]{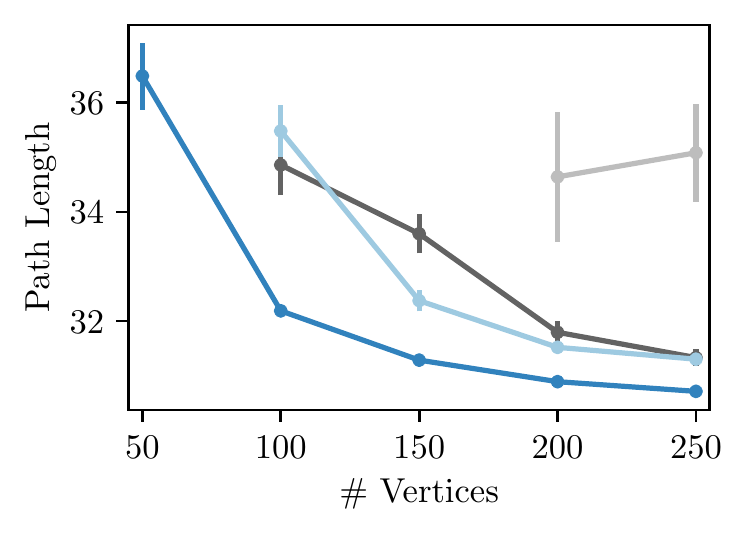}
\caption{Mean solution path lengths.}
\label{fig:num_samples_vs_solved_path_lengths}
\end{subfigure}
\caption{
  \intel environment with 100 random trials.
  \subref{fig:num_samples_vs_percent_solved}
  SV-PRM(Uniform+Rejection) succeeds on
  over 40\% with 50 vertices
  and all 100 trials with 100 vertices.
  \subref{fig:num_samples_vs_solved_path_lengths}
  SV-PRM(Uniform+Rejection) has lower solution path length than both PRM baselines, across all roadmap sizes.
  SV-PRM(Uniform) achieves comparable average solution path length to PRM(Uniform+Rejection), while solving a higher rate of problems.
}\vspace{-5pt}
\label{fig:bhm_query_dependent}
\end{figure}

\Figref{fig:coverage} demonstrates that SV-PRMs improve the configuration space coverage relative to the corresponding random initialization.
Note that this coverage metric does not characterize the connectivity of the roadmap;
as demonstrated in \Tabref{tab:toy2dqual},
rejection sampling can achieve good coverage without capturing the connectivity of the underlying space and yielding a feasible path. On query-dependent metrics,
\Figref{fig:bhm_query_dependent}
shows that both SV-PRM initializations enjoy shorter or comparable paths for the same number of vertices as the PRM baselines,
while solving a higher rate of problems.

\begin{figure}[!t]
\centering
\includegraphics[width=\linewidth]{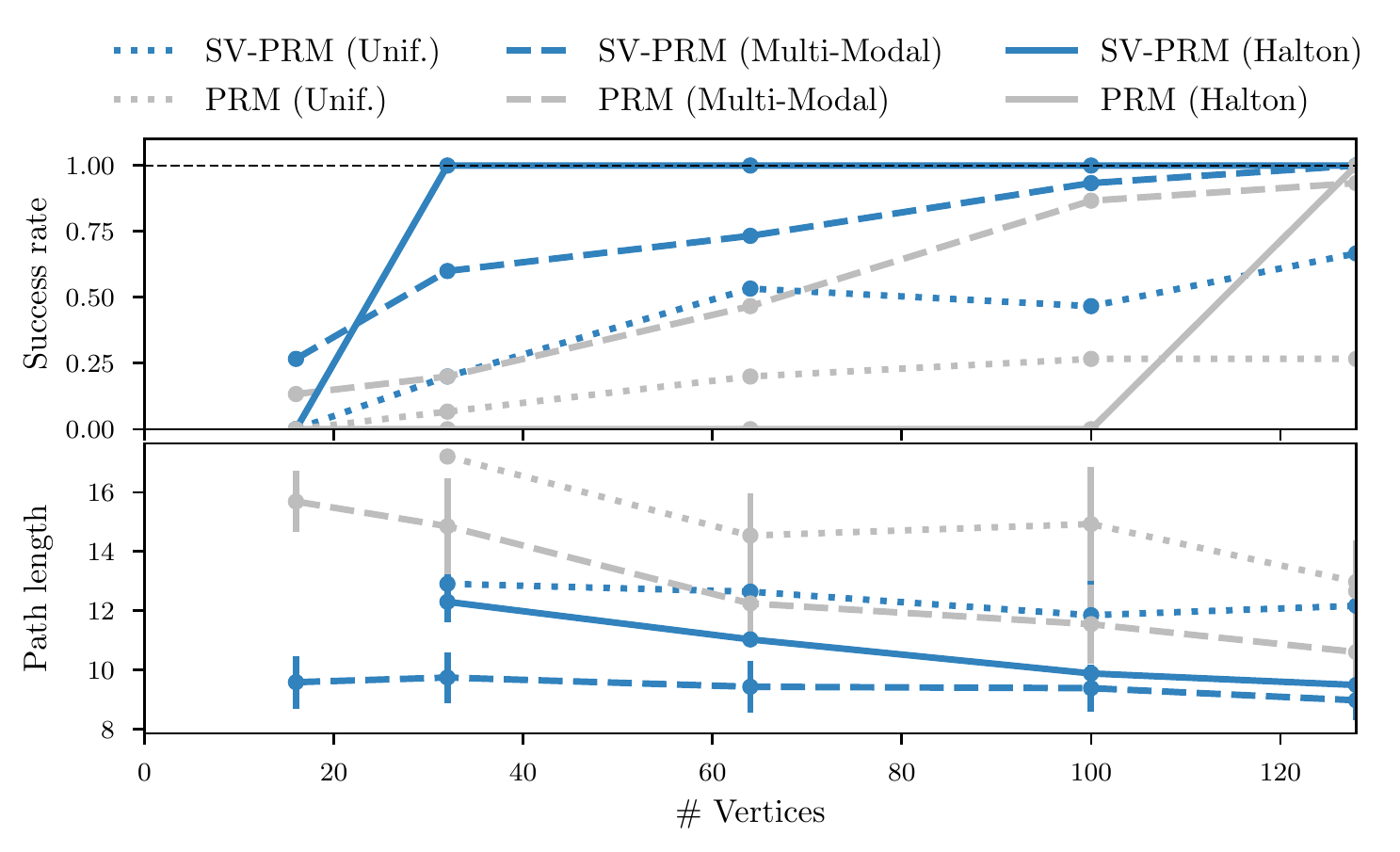}
\caption{ Solution statistics on the manipulation reaching experiment corresponding to \cref{subfig:main_franka}. Success rate and average path length of solutions are compared between SV-PRM and the PRM baseline for 30 trials (10 per random seed) using different initialization distributions (\ie~priors) across varying number of vertices: $N\in\{16, 32, 64, 100, 128\}$. Particle-based posterior sampling with SV-PRM significantly improves the planning solution, particularly in the regime of few vertices.}
\label{fig:franka_statistics}
\vspace{-1\baselineskip}\vspace{-5pt}
\end{figure}

Results for the 7-dof \franka task are shown in  \cref{fig:franka_statistics}, where we compare path statistics between SV-PRM and PRM for random and quasi-random initializations.  SV-PRM manages to generate a higher degree of feasible joint-space across vertex counts for all cases, and yields lower path costs when solutions are found. As can be seen in the accompanying video, performing inference with SVGD before graph construction pushes pose configurations into collision-free regions, increasing coverage in feasible space.


\section{Discussion and Future Work}
\label{sec:discussion}

The Stein Variational Probabilistic Roadmap proves to be a robust method for efficiently generating graphs well-suited for multi-query planning, outperforming existing biased-sampler PRM approaches.
Future work will investigate how to pick the appropriate prior distributions as well as number of particles, with more emphasis on planning for manipulation as these results were very promising.
Another exciting result worth investigating further is the tunable optimism parameter for partially explored environments.
This is especially applicable when running the algorithm in an online setting where the particles are updated incrementally as new observations arrive, similar in spirit to the results in \cite{senanayake2017bhm}.
Finally,
this approach would be an excellent complement to learned neural network samplers that leverage experience to guide sampling~\cite{ichter2018learning,kumar2019lego,qureshi2021mpnet,ichter2020critical},
where SV-PRMs can continue to optimize these biased samples
or directly incorporate the scoring functions as a probabilistic cost.



\clearpage  

\balance
\bibliographystyle{IEEEtran}
\bibliography{references}

\begin{thebibliography}{10}
\providecommand{\url}[1]{#1}
\csname url@rmstyle\endcsname
\providecommand{\newblock}{\relax}
\providecommand{\bibinfo}[2]{#2}
\providecommand\BIBentrySTDinterwordspacing{\spaceskip=0pt\relax}
\providecommand\BIBentryALTinterwordstretchfactor{4}
\providecommand\BIBentryALTinterwordspacing{\spaceskip=\fontdimen2\font plus
\BIBentryALTinterwordstretchfactor\fontdimen3\font minus
  \fontdimen4\font\relax}
\providecommand\BIBforeignlanguage[2]{{%
\expandafter\ifx\csname l@#1\endcsname\relax
\typeout{** WARNING: IEEEtran.bst: No hyphenation pattern has been}%
\typeout{** loaded for the language `#1'. Using the pattern for}%
\typeout{** the default language instead.}%
\else
\language=\csname l@#1\endcsname
\fi
#2}}

\bibitem{kavraki1996prm}
L.~E. Kavraki, P.~\v{S}vestka, J.-C. Latombe, and M.~H. Overmars,
  ``Probabilistic roadmaps for path planning in high-dimensional configuration
  spaces,'' \emph{{IEEE} Transactions on Robotics and Automation}, vol.~12,
  no.~4, pp. 566--580, 1996.

\bibitem{karaman2011sbmp}
S.~Karaman and E.~Frazzoli, ``Sampling-based algorithms for optimal motion
  planning,'' \emph{The International Journal of Robotics Research}, vol.~30,
  no.~7, pp. 846--894, 2011.

\bibitem{solovey2018rgg}
K.~Solovey, O.~Salzman, and D.~Halperin, ``New perspective on sampling-based
  motion planning via random geometric graphs,'' \emph{The International
  Journal of Robotics Research}, vol.~37, no.~10, pp. 1117--1133, 2018.

\bibitem{janson2018deterministic}
L.~Janson, B.~Ichter, and M.~Pavone, ``Deterministic sampling-based motion
  planning: Optimality, complexity, and performance,'' \emph{The International
  Journal of Robotics Research}, vol.~37, no.~1, pp. 46--61, 2018.

\bibitem{hsu1997expansive}
D.~Hsu, J.-C. Latombe, and R.~Motwani, ``Path planning in expansive
  configuration spaces,'' in \emph{{IEEE} International Conference on Robotics
  and Automation}, 1997.

\bibitem{liu2016svgd}
Q.~Liu and D.~Wang, ``{Stein} variational gradient descent: A general purpose
  {Bayesian} inference algorithm,'' in \emph{Advances in Neural Information
  Processing Systems}, 2016.

\bibitem{ocallaghan2012gpom}
S.~T. O'Callaghan and F.~T. Ramos, ``Gaussian process occupancy maps,''
  \emph{The International Journal of Robotics Research}, vol.~31, no.~1, pp.
  42--62, 2012.

\bibitem{ramos2016hilbert}
F.~Ramos and L.~Ott, ``Hilbert maps: Scalable continuous occupancy mapping with
  stochastic gradient descent,'' \emph{The International Journal of Robotics
  Research}, vol.~35, no.~14, pp. 1717--1730, 2016.

\bibitem{senanayake2017bhm}
R.~Senanayake and F.~Ramos, ``{Bayesian Hilbert Maps} for dynamic continuous
  occupancy mapping,'' in \emph{Conference on Robot Learning}, 2017.

\bibitem{ratliff2009chomp}
N.~Ratliff, M.~Zucker, J.~A. Bagnell, and S.~S. Srinivasa, ``{CHOMP}: Gradient
  optimization techniques for efficient motion planning,'' in \emph{{IEEE}
  International Conference on Robotics and Automation}, 2009.

\bibitem{zucker2013chomp}
M.~Zucker, N.~Ratliff, A.~D. Dragan, M.~Pivtoraiko, M.~Klingensmith, C.~M.
  Dellin, J.~A. Bagnell, and S.~S. Srinivasa, ``{CHOMP}: Covariant hamiltonian
  optimization for motion planning,'' \emph{The International Journal of
  Robotics Research}, vol.~32, no. 9-10, pp. 1164--1193, 2013.

\bibitem{mukadam2016gpmp}
M.~Mukadam, X.~Yan, and B.~Boots, ``Gaussian process motion planning,'' in
  \emph{{IEEE} International Conference on Robotics and Automation}, 2016.

\bibitem{mukadam2018gpmp}
M.~Mukadam, J.~Dong, X.~Yan, F.~Dellaert, and B.~Boots, ``Continuous-time
  {Gaussian} process motion planning via probabilistic inference,'' \emph{The
  International Journal of Robotics Research}, vol.~37, no.~11, pp. 1319--1340,
  2018.

\bibitem{boor1999gaussian}
V.~Boor, M.~H. Overmars, and A.~F. van~der Stappen, ``The {Gaussian} sampling
  strategy for probabilistic roadmap planners,'' in \emph{{IEEE} International
  Conference on Robotics and Automation}, 1999.

\bibitem{holleman2000medial}
C.~Holleman and L.~E. Kavraki, ``A framework for using the workspace medial
  axis in {PRM} planners,'' in \emph{{IEEE} International Conference on
  Robotics and Automation}, 2000.

\bibitem{hsu2003bridge}
D.~Hsu, T.~Jiang, J.~Reif, and Z.~Sun, ``The bridge test for sampling narrow
  passages with probabilistic roadmap planners,'' in \emph{{IEEE} International
  Conference on Robotics and Automation}, 2003.

\bibitem{saroya2021neuralgas}
M.~Saroya, G.~Best, and G.~A. Hollinger, ``Roadmap learning for probabilistic
  occupancy maps with topology-informed growing neural gas,'' \emph{{IEEE}
  Robotics and Automation Letters}, vol.~6, no.~3, pp. 4805--4812, 2021.

\bibitem{ichter2018learning}
B.~Ichter, J.~Harrison, and M.~Pavone, ``Learning sampling distributions for
  robot motion planning,'' in \emph{{IEEE} International Conference on Robotics
  and Automation}, 2018.

\bibitem{qureshi2021mpnet}
A.~H. Qureshi, Y.~Miao, A.~Simeonov, and M.~C. Yip, ``Motion planning networks:
  Bridging the gap between learning-based and classical motion planners,''
  \emph{{IEEE} Transactions on Robotics}, vol.~37, no.~1, pp. 48--66, 2021.

\bibitem{kumar2019lego}
R.~Kumar, A.~Mandalika, S.~Choudhury, and S.~S. Srinivasa, ``{LEGO}: Leveraging
  experience in roadmap generation for sampling-based planning,'' in
  \emph{{IEEE/RSJ} International Conference on Intelligent Robots and Systems},
  2019.

\bibitem{ichter2020critical}
B.~Ichter, E.~Schmerling, T.-W.~E. Lee, and A.~Faust, ``Learned critical
  probabilistic roadmaps for robotic motion planning,'' in \emph{{IEEE}
  International Conference on Robotics and Automation}, 2020.

\bibitem{lambert_stein_2020}
A.~Lambert, A.~Fishman, D.~Fox, B.~Boots, and F.~Ramos, ``Stein variational
  model predictive control,'' in \emph{Proceedings of the 4th Annual Conference
  on Robot Learning}.

\bibitem{barcelos2021dual}
L.~Barcelos, A.~Lambert, R.~Oliveira, P.~Borges, B.~Boots, and F.~Ramos, ``Dual
  online stein variational inference for control and dynamics,'' in
  \emph{Robotics: Science and Systems}, 2021.

\bibitem{lambert2021entropy}
A.~Lambert and B.~Boots, ``Entropy regularized motion planning via stein
  variational inference,'' \emph{RSS Workshop on Integrating Planning and
  Learning}, 2021.

\bibitem{thrun2005probabilistic}
S.~Thrun, W.~Burgard, and D.~Fox, \emph{Probabilistic Robotics}.\hskip 1em plus
  0.5em minus 0.4em\relax MIT Press, 2005.

\bibitem{schulman2014trajopt}
J.~Schulman, Y.~Duan, J.~Ho, A.~Lee, I.~Awwal, H.~Bradlow, J.~Pan, S.~Patil,
  K.~Goldberg, and P.~Abbeel, ``Motion planning with sequential convex
  optimization and convex collision checking,'' \emph{The International Journal
  of Robotics Research}, vol.~33, no.~9, pp. 1251--1270, 2014.

\bibitem{wang2019matrixsvgd}
D.~Wang, Z.~Tang, C.~Bajaj, and Q.~Liu, ``{Stein} variational gradient descent
  with matrix-valued kernels,'' in \emph{Advances in Neural Information
  Processing Systems}, 2019.

\bibitem{chen2019steinpm}
W.~Y. Chen, A.~Barp, F.-X. Briol, J.~Gorham, M.~Girolami, L.~Mackey, and
  C.~Oates, ``{Stein Point Markov Chain Monte Carlo},'' in \emph{International
  Conference on Machine Learning}, 2019.

\bibitem{detommaso2018svn}
G.~Detommaso, T.~Cui, A.~Spantini, Y.~Marzouk, and R.~Scheichl, ``A {Stein}
  variational {Newton} method,'' in \emph{Advances in Neural Information
  Processing Systems}, 2018.

\bibitem{dellin2016lazysp}
C.~Dellin and S.~S. Srinivasa, ``A unifying formalism for shortest path
  problems with expensive edge evaluations via lazy best-first search over
  paths with edge selectors,'' in \emph{International Conference on Automated
  Planning and Scheduling}, 2016.

\bibitem{Radish}
\BIBentryALTinterwordspacing
A.~Howard and N.~Roy, ``The robotics data set repository (radish),'' 2003.
  [Online]. Available: \url{http://radish.sourceforge.net/}
\BIBentrySTDinterwordspacing

\bibitem{halton1960sequence}
J.~H. Halton, ``On the efficiency of certain quasi-random sequences of points
  in evaluating multi-dimensional integrals,'' \emph{Numerische Mathematik},
  vol.~2, no.~1, pp. 84--90, 1960.

\bibitem{gretton2008mmd}
A.~Gretton, K.~Borgwardt, M.~J. Rasch, B.~Scholkopf, and A.~J. Smola, ``A
  kernel method for the two-sample problem,'' in \emph{Advances in Neural
  Information Processing Systems}, 2008.

\end{thebibliography}

\end{document}